\documentclass{article}

\usepackage[nonatbib,final]{nips_2017}

\usepackage[utf8]{inputenc} 
\usepackage[T1]{fontenc}    
\usepackage{hyperref}       
\usepackage{url}            
\usepackage{booktabs}       
\usepackage{amsfonts}       
\usepackage{amsmath}
\usepackage{amsthm}
\usepackage{amssymb}
\usepackage{nicefrac}       
\usepackage{microtype}      
\usepackage{graphicx}
\usepackage{caption}
\usepackage{subcaption}

\title{Deep Text-to-Speech System with Seq2Seq Model}

\author{
	Yuan (Gary) Wang \\
	Computer Science Department\\
	Simon Fraser University\\
	Burnaby, BC \\
	\texttt{ywa289@sfu.ca} \\
}

\begin{document}	
	\maketitle
	\begin{abstract}
		Recent trends in neural network based text-to-speech/speech synthesis pipelines have employed recurrent Seq2seq architectures that can synthesize realisitic sounding speech directly from text characters. These systems however have complex architectures and takes a substantial amount of time to train.We introduce several modifications to these Seq2seq architectures that allow for faster training time, and also allows us to reduce the complexity of the model architecture at the same time. We show that our proposed model can achieve attention alignment much faster than previous architectures and that good audio quality can be achieved with a model that's much smaller in size. Sample audio available at \textbf{\url{https://soundcloud.com/gary-wang-23/sets/tts-samples-for-cmpt-419}}
	\end{abstract}
	
	\section{Introduction}
	\label{sec:intro}
	
	Traditional text-to-speech (TTS) systems are composed of complex pipelines \cite{Taylor09:TTS-Review}, these often include accoustic frontends, duration model, acoustic prediction model and vocoder models. The complexity of the TTS problem coupled with the requirement for deep domain expertise means these systems are often brittle in design and results in un-natural synthesized speech.
	
	The recent push to utilize deep, end-to-end TTS architectures \cite{Wang17:Tacotron-Original}\cite{Arik17:Deepvoice-1} that can be trained on <text,audio> pairs shows that deep neural networks can indeed be used to synthesize realistic sounding speech, while at the same time eliminating the need for complex sub-systems that neede to be developed and trained seperately.
	
	The problem of TTS can be summed up as a signal-inversion problem: given a highly compressed source signal (text), we need to invert or "decompress" it into audio. This is a difficult problem as there're multi ways for the same text to be spoken. In addtion, unlike end-to-end translation or speech recognition, TTS ouptuts are continuous, and output sequences are much longer than input squences.
	
	Recent work on neural TTS can be split into two camps, in one camp Seq2Seq models with recurrent architectures are used \cite{Wang17:Tacotron-Original}\cite{Sotelo17:char2wav}. In the other camp, full convolutional Seq2Seq models are used \cite{Arik17:Deepvoice-1}. Our model belongs in the first of these classes using recurrent architectures. Specifically we make the following contributions:
\begin{enumerate}
\item We propose replacing location sensitive + additive attention from previous work \cite{Shen18:Tacotron-2} with a much simpler query-key attention (attention mechanism similar to \cite{Vaswani:Transformer})
\item We propose adding a monotonic attention loss (similar to \cite{Tachibana18:DCTTS}) that forces the attention to correctly align much earlier in training
\end{enumerate}
	
	\section{Related Work}
	\label{sec:related}
	Neural text-to-speech systems have garnered large research interest in the past 2 years. The first to fully explore this avenue of research was Google's tacotron \cite{Wang17:Tacotron-Original} system. Their architecture based off the original Seq2Seq framework. In addition to encoder/decoder RNNs from the original Seq2Seq , they also included a bottleneck prenet module termed CBHG, which is composed of sets of 1-D convolution networks followed by highway residual layers. The attention mechanism follows the original Seq2Seq \cite{Bahdanau15:OG-Seq2Seq} mechanism (often termed Bahdanau attention). This is the first work to propose training a Seq2Seq model to convert text to mel spectrogram, which can then be converted to audio wav via iterative algorithms such as Griffin Lim \cite{Griffin84:griffin-lim}.
	
	A parrallel work exploring Seq2Seq RNN architecture for text-to-speech was called Char2Wav \cite{Sotelo17:char2wav}. This work utilized a very similar RNN-based Seq2Seq architecture, albeit without any prenet modules. The attention mechanism is guassian mixture model (GMM) attention from Alex Grave's work. Their model mapped text sequence to 80 dimension vectors used for the WORLD Vocoder \cite{Morise:world-vocoder}, which invert these vectors into audio wave.
	
	More recently, a fully convolutional Seq2Seq architecture was investigated by Baidu Research \cite{Arik17:Deepvoice-1} \cite{Ping17:deepoice-3}. The deepvoice architecture is composed of causal 1-D convolution layers for both encoder and decoder. They utilized query-key attention similar to that from the transformer architecure \cite{Vaswani:Transformer}.
	
	Another fully convolutional Seq2Seq architecture known as DCTTS was proposed \cite{Tachibana18:DCTTS}. In this architecture they employ modules composed of Causal 1-D convolution layers combined with Highway networks. In addition they introduced methods for help guide attention alignments early. As well as a forced incremental attention mechanism that ensures monotonic increasing of attention read as the model decodes during inference.

\section{Model Overview}
\label{sec:model-architecture}

The architecture of our model utilizes RNN-based Seq2Seq model for generating mel spectrogram from text. The architecture is similar to that of Tacotron 2 \cite{Shen18:Tacotron-2}. The generated mel spetrogram can either be inverted via iterative algorithms such as Griffin Lim, or through more complicated neural vocoder networks such as a mel spectrogram conditioned Wavenet \cite{Oord16:wavenet-1}.

Figure \ref{fig:architecture} below shows the overall architecture of our model.

\begin{figure}[!h]
		\centering
		\includegraphics[width=120mm]{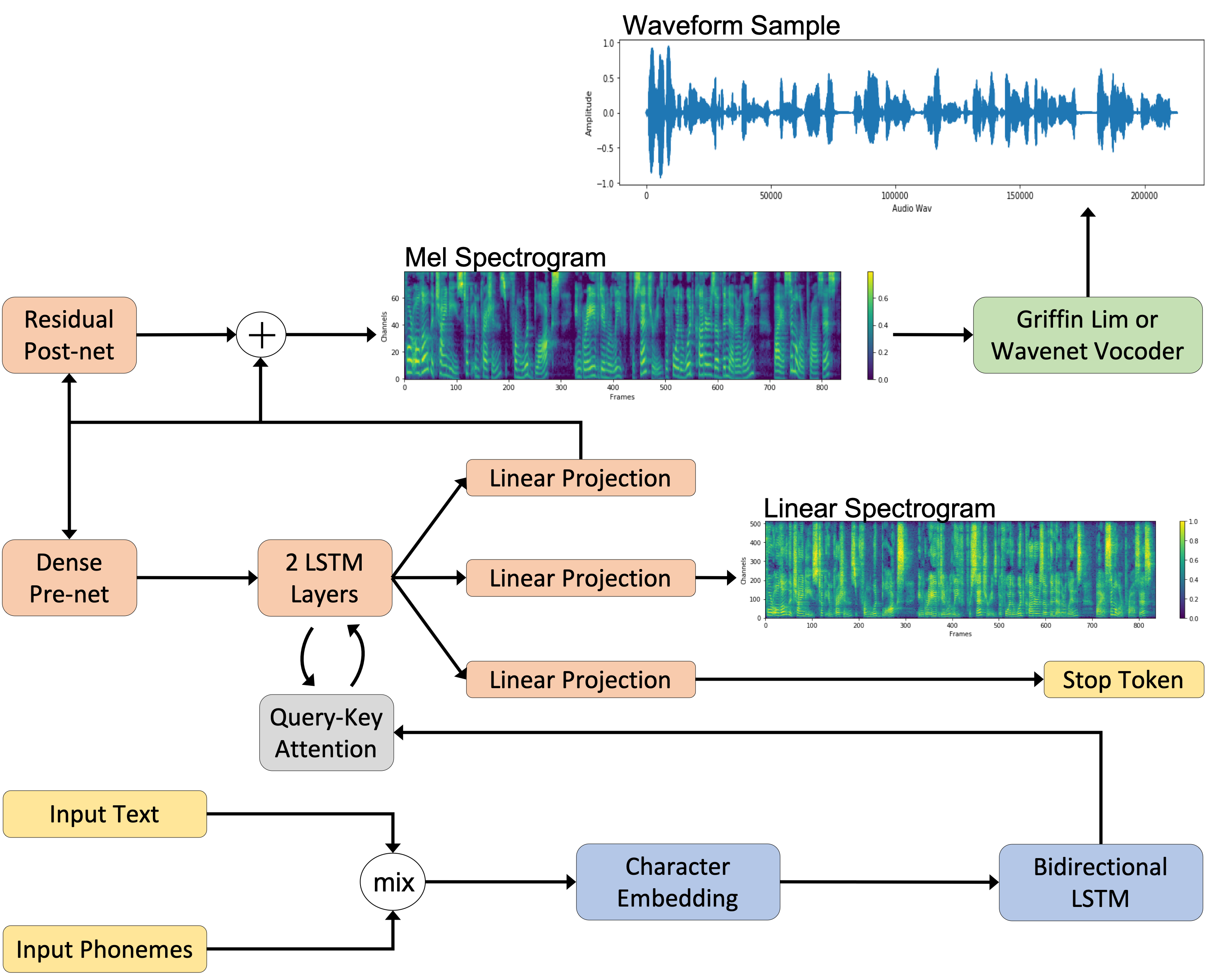}
		\caption{Overall architecture of our Seq2Seq model for neural text-to-speech. Note that inputs, encoder, decoder and attention are labelled different colors.}
		\label{fig:architecture}
	\end{figure}

\subsection{Text Encoder}
The encoder acts to encoder the input text sequence into a compact hidden representation, which is consumed by the decoder at every decoding step. The encoder is composed of a $d$-dim embedding layer that maps the input sequence into a dense vector. This is followed by a 1-layer bidirectional LSTM/GRU with $d$ hidden dim ($2\times d$ hidden dim total for both directions). two linear projections layers project the LSTM/GRU hidden output into two vectors $K$ and $V$ of the same $d$-dimension, these are the \textbf{key} and \textbf{value} vectors.

\begin{align}
L = [l_1, l_2, ... l_N] \in Char^N \\
(K, V) = TextEncoder(L)
\end{align}

where $K, V \in \mathbb{R}^{d \times N}$.

\subsection{Query-Key Attention}
Query key attention is similar to that from transformers \cite{Vaswani:Transformer}. Given $K$ and $V$ from the encoder, the query, $Q \in \mathbb{R}^{d \times 1}$, is computed from a linear transform of the concatenation of previous decoder-rnn hidden state, $dh_{t-1} \in \mathbb{R}^d$, combined with attention-rnn hidden state, $ah_{t-1} \in \mathbb{R}^d$).
\begin{equation}
Q = Linear([dh_{t-1}, ah_{t-1}])
\end{equation}

Given $Q$, $K$, $V$, the attention at each decoding step is computed by the scaled dot-product operation as:
\begin{equation}
Attetion(Q, K, V) = softmax(\frac{QK^T}{\sqrt{d}})V
\end{equation}

Note that similar to transformers \cite{Vaswani:Transformer}, we apply a scale the dot-product by $d$ to prevent softmax function into regions where it has extremely small gradients.

\subsection{Decoder}
The decoder is an autoregressive recurrent neural network that predicts mel spectrogram from the encoded input sentence one frame at a time.

\subsubsection{Decoder RNNs}
The decoder decodes the hidden representation from the encoder, with the guidance of attention. The decoder is composed of two uni-directional LSTM/GRU with $d$ hidden dimensions. The first LSTM/GRU, called the AttentionRNN, is for computing attention-mechanism related items such as the attention query $Q$.
\begin{equation}
ah_t = AttentionRNN([Q_{t-1}, ah_{t-1}])
\end{equation}

The second LSTM/GRU, DecoderRNN, is used to compute the decoder hidden output, $dh_t$.

\begin{equation}
dh_t = DecoderRNN([Q_{t-1}, dh_{t-1}])
\end{equation}

\subsubsection{Input Mel Spectrogram Prenet}
A 2-layer dense prenet of dimensions (256,256) projects the previous mel spectrogram output $M_{t-1} \in \mathbb{R}^{80 \times 1}$ into hidden dimension $d$. Similar to Tacotron 2, the prenet acts as an information bottleneck to help produce useful representation for the downstream attention mechanism. Our model differs from Tacotron 2 in that we jointly project 5 consequetive mel frames at once into our hidden representation, which is faster and unlike Tacotron 2 which project 1 mel frame at at time.

\subsubsection{Mel Spectrogram Projection}
The DecoderRNN's hidden state $dh_t$ is also projected to mel spectrogram $M' \in \mathbb{R}^{80 \times 1}$. A residual post-net composed of 2 dense layer followed by a \textit{tanh} activation function also projects the same decoder hidden state $dh_t$ to mel spectrogram $M^r \in \mathbb{R}^{80 \times 1}$, which is added to the linear projected mel $M'$ to produce the final mel spectrogram $M$.

\begin{align}
M'_t = Linear(dh_t)\\
M^r_t = PostNet(dh_t)\\
M_t = M'_t + M^r_t
\end{align}

\subsubsection{Linear Spectrogram Projection}
A linear spectrogram $S \in \mathbb{R}^{513 \times 1}$ is also computed from a linear projection of the decoder hidden state $dh_t$. This acts as an additional condition on the decoder hidden input.
\begin{equation}
S_t = Linear(dh_t)
\end{equation}

\subsubsection{Stop Token Prediction}
A single scalar stop token is computed from a linear projection of the decoder hidden state $dh_t$ to a scalar, followed by $\sigma$, or sigmoid function. This stop token allows the model to learn when to stop decoding during inference. During inference, if stop token output is $> 0.5$, we stop decoding.
\begin{equation}
STOP_t = \sigma(Linear(dh_t))
\end{equation}

\subsection{Training and Loss}
Total loss on the model is computed as the sum of 3 component losses: 1. Mean-Squared-Error(MSE) of predicted and ground-truth mel spectrogram 2. MSE of Linear Spectrogram 3. Binary Cross Entropy Loss of our stop token. Adam optimizer is used to optimize the model with learning rate of $1e-4$.

Model is trained via \textbf{teacher forcing}, where the ground-truth mel spectrogram is supplied at every decoding step instead of the model's own predicted mel spectrogram. To ensure the model can learn for long term sequences, teacher forcing ratio is annealed from 1.0 (full teacher forcing) to 0.2 (20 percent teacher forcing) over 300 epochs.

\section{Proposed Improvements}

Our proposed improvements come from the observation that employing generic Seq2seq models for TTS application misses out on further optimization that can be achieved when we consider the specific problem of TTS. Specifically, we notice that in TTS, unlike in applications like machine translation, the Seq2Seq attention mechanism should be mostly monotonic. In other words, when one reads a sequence of text, it is natural to assume that the text position progress nearly linearly in time with the sequence of output mel spectrogram. With this insight, we can make 3 modifications to the model that allows us to train faster while using a a smaller model. 

\subsection{Changes to Attention Mechanism}
In the original Tacotron 2, the attention mechanism used was location sensitive attention \cite{Chorowski15:location-attention} combined the original additive Seq2Seq \cite{Bahdanau15:OG-Seq2Seq} Bahdanau attention.

We propose to replace this attention with the simpler query-key attention from transformer model. As mentioned earlier, since for TTS the attention mechanism is an easier problem than say machine translation, we employ query-key attention as it's simple to implement and requires less parameters than the original Bahdanau attention.

\subsection{Guided Attention Mask}

Following the logic above, we utilize a similar method from \cite{Tachibana18:DCTTS} that adds an additional guided attention loss to the overall loss objective, which acts to help the attention mechanism become monotoic as early as possible.

As seen from \ref{fig:attention-mask}, an attention loss mask, $W_{n,t}$, is created applies a loss to force the attention alignment, $A_{n,t}$, to be nearly diagonal. That is:
\begin{equation}
L_{attn}(A) = \mathbb{E}_{nt}[A_{n,t} W_{n,t}]
\end{equation}

Where $W_{nt} = 1-\exp{-(n/N - t/T)^2/2g^2}$, $n$ is the $n$-th character, $N$ is the max character length, $t$ is the $t$-th mel frame, $T$ is the max mel frame, and $g$ is set at 0.2.
This modification dramatically speed up the attention alignment and model convergence.

\begin{figure}[!h]
		\centering
		\includegraphics[width=60mm]{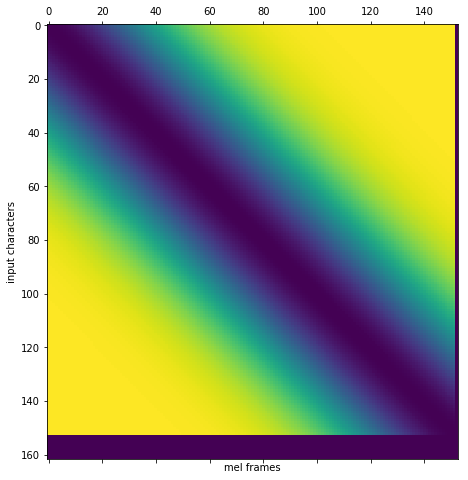}
		\caption{Attention guide mask. Note that bright area has larger values and dark area has small values.}
		\label{fig:attention-mask}
	\end{figure}
	
Figure 3 below shows the results visually. The two images are side by side comparison of the model's attention after 10k training steps. The image on the left is trained with the atention mask, and the image on the right is not. We can see that with the attention mask, clear attention alignment is achieved much faster.

\begin{figure}[!h]
		\centering
		\includegraphics[width=130mm]{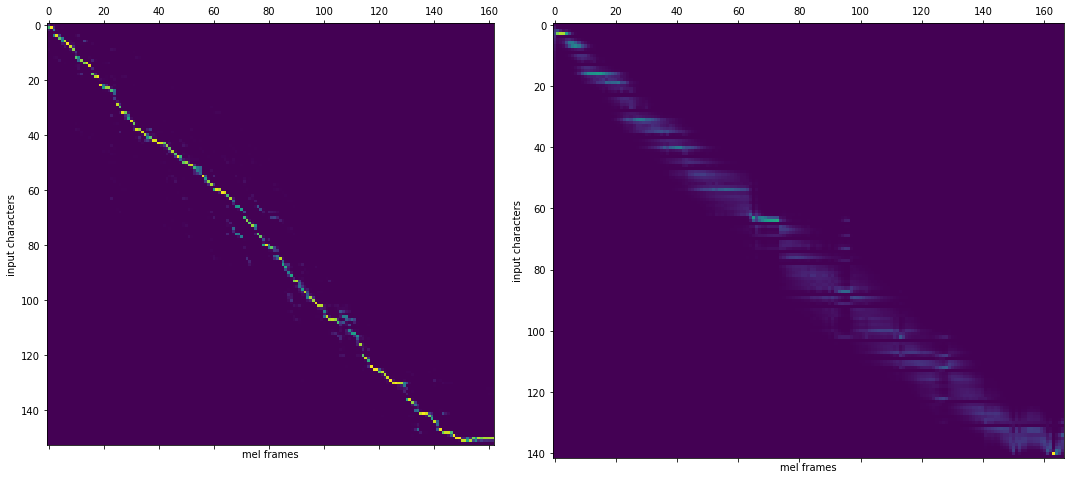}
		\caption{Attention alignment plots for two identifical models trained with and without guided attention masks. Both models have been trained for 10k in this figure.}
		\label{fig:attention-mask}
	\end{figure}
	
\subsection{Forced Incremental Attention}
During inference, the attention $A_{n,t}$ occasionally skips multiple charaters or stall on the same character for multiple output frames. To make generation more robust, we modify $A_{nt}$ during inference to force it to be diagonal.

The Forced incremental attention is implemented as follows:

Given $n_t$, the position of character read at $t$-th time frame, where $n_t = argmax_n A_{n,t}$, if $ 1 \leq n_t - n_{t-1} \leq 3$, the current attention is forcibly set to $A_{n,t} = \delta_{n,n_{t-1}+1}$, so that attention is incremental, i.e $n_t - n_{t-1}=1$.

\section{Experiment and Results}
\subsection{Experiment Dataset}
The open source LJSpeech Dataset was used to train our TTS model. This dataset contains around 13k <text,audio> pairs of a single female english speaker collect from across 7 different non-fictional books. The total training data time is around 21 hours of audio.

One thing to note that since this is open-source audio recorded in a semi-professional setting, the audio quality is not as good as that of proprietary internal data from Google or Baidu. As most things with deep learning, the better the data, the better the model and results.

\subsection{Experiment Procedure}
Our model was trained for 300 epochs, with batch size of 32. We used pre-trained opensource implementation of Tactron 2 (https://github.com/NVIDIA/tacotron2) as baseline comparison. Note this open-source version is trained for much longer (around 1000 epochs) however due to our limited compute we only trained our model up to 300 epochs

\subsection{Evaluation Metrics}
We decide to evaluate our model against previous baselines on two fronts, Mean Opnion Score (MOS) and training speed. 

Typical TTS system evaluation is done with mean opinion score (MOS). To compute this score, many samples of a TTS system is given to human evaluators and rated on a score from 1 (Bad) to 5 (Excellent). the MOS is then computed as the arithmetic mean of these score:
\begin{equation}
MOS = \frac{\sum^N_{n-1}}R_n{N}
\end{equation}
Where $R$ are individual ratings for a given sample by N subjects.

For TTS models from google and Baidu, they utilized Amazon mechanical Turk to collect and generate MOS score from larger number of workers. However due to our limited resources, we chose to collect MOS score from friends and families (total 6 people).

For training time comparison, we choose the training time as when attention alignment start to become linear and clear. After digging through the git issues in the Tacotron 2 open-source implementation, we found a few posts where users posted their training curve and attention alignment during training (they also used the default batch size of 32). We used their training steps to roughly estimate the training time of Tacotron 2 when attention roughly aligns. For all other models the training time is not comparable as they either don't apply (e.g parametric model) or are not reported (Tacotron griffin lim, Deepvoice 3).

	\begin{table}[!h]
		\centering
		\begin{tabular}{|l|l|l|}
			\hline
			\textbf{TTS Models} & \textbf{MOS Score}  & \textbf{Training Time (h)}
			 \\ \hline
			Parametric   & $3.492 \pm 0.096$ & N/A \\ \hline
			Tacotron (Griffin Lim)  & $4.001 \pm 0.087$ & N/A \\ \hline
			Concatenative       & $4.166 \pm 0.091$ & N/A \\ \hline
			Deepvoice 3       & $3.780 \pm 0.034$ & N/A \\ \hline		
			Tacotron 2 (Open source)  & $3.528 \pm 0.179$ & 15 \\ \hline
			\textbf{Our Model} & $3.440 \pm 0.182$ & 3 \\ \hline
		\end{tabular}
		\caption{Result table comparing MOS score and rough estimated training time between different TTS systems.}
		\label{ta:smallflop}
		\vspace{-0.1in}
	\end{table}

Direct comparison of model parameters between ours and the open-source tacotron 2, our model contains 4.5 million parameters, whereas the Tacotron 2 contains around 13 million parameters with default setting. By helping our model learn attention alignment faster, we can afford to use a smaller overall model to achieve similar quality speech quality.

	\section{Conclusion}
	We introduce a new architecture for end-to-end neural text-to-speech system. Our model relies on RNN-based Seq2seq architecture with a query-key attention. We introduce novel guided attention mask to improve model training speed, and at the same time is able to reduce model parameters. This allows our model to achieve attention alignment at least 3 times faster than previous RNN-based Seq2seq models such as Tacotron 2. We also introduce forced incremental attention during synthesis to prevent attention alignment mistakes and allow model to generate coherent speech for very long sentences.

	\bibliographystyle{plain}

\end{document}